\documentclass[11pt]{article}

\usepackage[preprint]{acl}

\usepackage{times}
\usepackage{latexsym}

\usepackage[T1]{fontenc}

\usepackage[utf8]{inputenc}

\usepackage{microtype}

\usepackage{inconsolata}

\usepackage{graphicx}

\usepackage{CJKutf8}
\usepackage{tikz}
\usepackage{color}
\usepackage{xcolor}
\usepackage{tabularx}
\usepackage[most]{tcolorbox}
\usepackage{pifont}
\usepackage{amssymb}
\usepackage{amsmath}
\usepackage{enumerate}
\usepackage{enumitem}
\usepackage{afterpage}
\usepackage{multirow}
\usepackage{rotating}
\usepackage{algorithm}
\usepackage{algpseudocode}

\usepackage{expex}

\lingset{everygla=\upshape}

\definecolor{darkgreen}{rgb}{0.00, 0.70, 0.12}

\title{Translation via Annotation: A Computational Study of Translating Classical Chinese into Japanese}

\author{Zilong Li \\
  Department of Linguistics\\
  University of Colorado, Boulder\\
  \texttt{\url{Zilong.Li@colorado.edu}} \\\And
  Jie Cao \\
  School of Computer Science\\
  University of Oklahoma\\
  \texttt{\url{jie.cao@ou.edu}} \\}

\begin{document}
\maketitle
\begin{abstract}
Ancient people translated classical Chinese into Japanese using a system of annotations placed around characters. We abstract this process as sequence tagging tasks and fit them into modern language technologies. The research on this annotation and translation system faces a low-resource problem. We alleviate this problem by introducing an LLM-based annotation pipeline and constructing a new dataset from digitized open-source translation data. We show that in the low-resource setting, introducing auxiliary Chinese NLP tasks enhances the training of sequence tagging tasks. We also evaluate the performance of Large Language Models (LLMs) on this task. While they achieve high scores on direct machine translation, our method could serve as a supplement to LLMs to improve the quality of character's annotation.\footnote{Our code and data are available at \url{https://github.com/shiryusann/KanbunKundoku}.}
\end{abstract}

\section{Introduction}

Classical Chinese (5\textsuperscript{th} century B.C.E. to 19\textsuperscript{th} century A.D.) has a long history in Japan. Although the exact timing of classical Chinese's introduction to Japan remains unclear, the presence of classical Chinese in Japan dates back to at least the 8\textsuperscript{th} century A.D. The two oldest classical Japanese books, \textit{Kojiki} and \textit{Nihon Shoki}, are written entirely in classical Chinese. In Japan, classical Chinese is referred to as \textit{Kanbun}. During the adoption of \textit{Kanbun}, Japanese people developed an annotation and reading system, called \textit{Kundoku}, by which they translated classical Chinese into Japanese. Today, this annotation system continues to play a role in education and humanities research. Understanding this annotation system helps us to build educational software to assist Japanese education. It can also benefit the digital humanities research of East Asian history and literature.

\begin{figure}[t]
    \centering
    \begin{minipage}[t]{0.30\columnwidth}
        \centering
        \includegraphics[scale = 0.50]{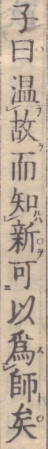}
    \end{minipage}
    \hspace{-1cm}
    \begin{minipage}[t]{0.70\columnwidth}
        \centering 
        \begin{tikzpicture}[x = 0.75pt, y = 0.75pt, yscale = -1, xscale = 1]
            \tikzset{every picture/.style={line width=0.75pt}}
            \draw (20, 0) node  [align = left, font = \small] {子};
            \draw (20, 15) node  [align = left, font = \small] {曰};
            \draw (35, 12) node  [align = left, font = \small] {\textcolor{darkgreen}{。}};
            \draw (20, 30) node  [align = left, font = \small] {溫};
            \draw (29, 38) node  [align = left, font = \tiny] {\textcolor{red}{テ}};
            \draw (11, 38) node  [align = left, font = \tiny] {\textcolor{blue}{レ}};
            \draw (20, 45) node  [align = left, font = \small] {故};
            \draw (29, 53) node  [align = left, font = \tiny] {\textcolor{red}{ヲ}};
            \draw (20, 60) node  [align = left, font = \small] {而};
            \draw (20, 75) node  [align = left, font = \small] {知};
            \draw (29, 83) node  [align = left, font = \tiny] {\textcolor{red}{ル}};
            \draw (11, 83) node  [align = left, font = \tiny] {\textcolor{blue}{レ}};
            \draw (29, 88) node  [align = left, font = \tiny] {\textcolor{red}{ハ}};
            \draw (20, 90) node  [align = left, font = \small] {新};
            \draw (29, 98) node  [align = left, font = \tiny] {\textcolor{red}{ヲ}};
            \draw (35, 88) node  [align = left, font = \small] {\textcolor{darkgreen}{。}};
            \draw (20, 105) node  [align = left, font = \small] {可};
            \draw (11, 113) node  [align = left, font = \tiny] {\textcolor{blue}{二}};
            \draw (20, 120) node  [align = left, font = \small] {以};
            \draw (20, 135) node  [align = left, font = \small] {爲};
            \draw (29, 143) node  [align = left, font = \tiny] {\textcolor{red}{ス}};
            \draw (11, 143) node  [align = left, font = \tiny] {\textcolor{blue}{一}};
            \draw (11, 147) node  [align = left, font = \tiny] {\textcolor{blue}{レ}};
            \draw (20, 150) node  [align = left, font = \small] {師};
            \draw (29, 158) node  [align = left, font = \tiny] {\textcolor{red}{ト}};
            \draw (20, 165) node  [align = left, font = \small] {矣};
            \draw (35, 162) node  [align = left, font = \small] {\textcolor{darkgreen}{。}};
            
            \draw    (40,90) -- (68,90);
            \draw [shift={(70,90)}, rotate = 180] [color={rgb, 255:red, 0; green, 0; blue, 0 }  ][line width=0.75]    (6.56,-1.97) .. controls (4.17,-0.84) and (1.99,-0.18) .. (0,0) .. controls (1.99,0.18) and (4.17,0.84) .. (6.56,1.97);

            \draw (100, 0) node  [align = left, font = \small] {子};
            \draw (100, 15) node  [align = left, font = \small] {曰};
            \draw (100, 30) node  [align = left, font = \small] {\textcolor{darkgreen}{、}};
            \draw (100, 45) node  [align = left, font = \small] {故};
            \draw (100, 60) node  [align = left, font = \small] {\textcolor{red}{を}};
            \draw (100, 75) node  [align = left, font = \small] {温};
            \draw (100, 90) node  [align = left, font = \small] {\textcolor{red}{て}};
            \draw (100, 105) node  [align = left, font = \small] {新};
            \draw (100, 120) node  [align = left, font = \small] {\textcolor{red}{を}};
            \draw (100, 135) node  [align = left, font = \small] {知};
            \draw (100, 150) node  [align = left, font = \small] {\textcolor{red}{る}};
            \draw (100, 165) node  [align = left, font = \small] {\textcolor{red}{は}};
            \draw (80, 0) node  [align = left, font = \small] {\textcolor{darkgreen}{、}};
            \draw (80, 15) node  [align = left, font = \small] {以};
            \draw (80, 30) node  [align = left, font = \small] {師};
            \draw (80, 45) node  [align = left, font = \small] {\textcolor{red}{と}};
            \draw (80, 60) node  [align = left, font = \small] {為};
            \draw (80, 75) node  [align = left, font = \small] {\textcolor{red}{す}};
            \draw (80, 90) node  [align = left, font = \small] {可};
            \draw (80, 105) node  [align = left, font = \small] {\textcolor{darkgreen}{。}};
        \end{tikzpicture}
    \end{minipage}
    \caption{Classical Chinese sentence with marks and its Japanese translation. {\textcolor{darkgreen}{Green}} punctuations are \textbf{Kutōten} to segment sentences. {\textcolor{blue}{Blue}} symbols are \textbf{Kaeriten} indicating the reading order. {\textcolor{red}{Red}} characters are \textbf{Okurigana} for grammatical and inflectional roles.}
    \label{fig1}
\end{figure}
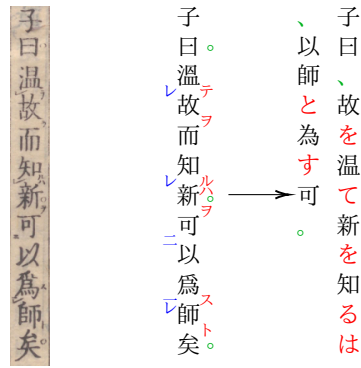

In the \textit{Kundoku} translation system, annotations are placed around each Chinese character using three types of marks: \textbf{Kutōten}, \textbf{Kaeriten}, and \textbf{Okurigana}, as illustrated in Figure~\ref{fig1}. These marks respectively indicate sentence punctuation, reading order, and grammatical/inflectional information, guiding readers to convert Classical Chinese text into coherent Japanese sentences.

\begin{CJK}{UTF8}{ipxm}
\end{CJK}

After annotation, people reconstruct Chinese characters with their marks. Chinese characters are first reordered according to the \textbf{Kaeriten} marks. Then, \textbf{Okurigana} marks are appended to the specific Chinese characters. Through this procedure, a Japanese translation of the given classical Chinese sentence is derived.

In the era of machine learning and Large Language Models (LLMs), researchers are increasingly applying modern language technologies to automate the annotation~\citep{yasuoka-2020-kanbun} and translation~\citep{wang-etal-2023-kanbun} of \textit{Kanbun}. However, there are two main challenges in developing automatic annotation and translation models for the \textit{Kundoku} system: (1) the lack of parallel corpus annotated with \textit{Kundoku} marks, and (2) the underexplored impact of classical Chinese linguistic knowledge on language modeling.

In this paper, we present a brief and systematic study of the \textit{Kundoku} annotation and translation system, focusing on addressing these two challenges. We focus exclusively on \textbf{Kaeriten} and \textbf{Okurigana}, since \textbf{Kutōten} (punctuations) are typically already provided in most classical Chinese books. Our contributions are as follows:

\begin{itemize}
    \item We theoretically analyze the expressiveness of \textbf{Kaeriten} marks and design an automaton with transducer that decodes characters with marks into Japanese sentences. This validates our hypothesis that reducing classical Chinese–Japanese translation to \textit{Kundoku} marks tagging tasks is both theoretically sound and practically feasible.
    \item To alleviate the challenge of the low-resource, we construct a new dataset from online digitized classical Chinese texts and their corresponding Japanese translations. We propose a LLM-based mark generation method, utilizing the automaton for validation to generate \textit{Kundoku} marks.
    \item To incorporate necessary classical Chinese knowledge, we fine-tune classical Chinese Language models on our new dataset using multi-task supervision with a set of auxiliary Chinese NLP tasks. Ablation studies on these auxiliary tasks provide an understanding of the role of classical Chinese knowledge in this task. Our optimal model outperforms previous baseline across many evaluation metrics, and even performs as comparable to some LLMs.
	\item We also conduct a comprehensive evaluation of several large language models (LLMs) on this task using both zero-shot and few-shot prompting, shedding light on future opportunities and challenges in applying LLM to this \textit{Kundoku} annotation and translation system.
\end{itemize}

\section{Related Works}

The task of translating classical Chinese into Japanese via language technologies was first proposed by~\citet{yasuoka-2018-reordering}. They designed a rule-based method leveraging Universal Dependencies (UD) parsing for classical Chinese, which generates \textbf{Kaeriten} marks based on part of speech (POS) tags and the direction and label of dependency arcs. Additionally, they defined specific rules to adjust the position of generated \textbf{Kaeriten} marks according to their context.~\citet{yasuoka-2020-kanbun} subsequently introduced a dictionary used to add \textbf{Okurigana} marks to Chinese characters. These two components together constitute a complete annotation system as described in the previous section.

\citet{wang-etal-2023-kanbun} built a pipeline to directly translate classical Chinese poems into Japanese using pretrained Language Models (PLMs). They decomposed the translation task into two stages: character ordering followed by text generation. In their pipeline, Chinese characters are first processed by BERT/RoBERTa models to determine their order within the target Japanese sentence. Then, the reordered characters are fed into mT5/mGPT models to generate the final translation. In addition to this pipeline, they released their dataset, comprising approximately 3,400 sentence pairs consisting of classical Chinese poems and the corresponding Japanese translations. However, it lacks annotations for \textbf{Kaeriten} and \textbf{Okurigana} marks.

Beyond research involving pretrained language models, other studies have explored the \textit{Kundoku} annotation system from the perspective of combinatorics and algorithms.~\citet{shimano-2009-permutation} investigated the expressiveness of \textbf{Kaeriten} marks. They proposed a tree-structure model and a matrix model to depict the reading process of Chinese characters. They further proposed a recursive formula that computes the number of character permutations that can be described by \textbf{Kaeriten}. In subsequent work,~\citet{shimano-2012-permutation} solved this recursive formula and derived the corresponding generating function. They summarized their models and demonstrated that the \textit{Kundoku} process is equivalent to a context-free language (CFL) with the Chomsky hierarchy~\citep{shimano-2018-formalization}.

In the domain of low-resource Japanese translation research,~\citet{mao-etal-2020-jass} incorporated Japanese syntactic knowledge into language models by introducing a word reordering training objective.~\citet{mao-etal-2022-ling} extended this strategy to more language pairs and achieved improvements in machine translation.

The translation of classical Chinese has also received attention in recent years. Existing research discusses different aspects of this field, including dataset construction (\citealp{wong-etal-2024-humanistic};~\citealp{liu-etal-2025-large}), evaluation (\citealp{zhou-etal-2023-wyweb};~\citealp{bennett-etal-2025-evaluating};~\citealp{chen-etal-2025-benchmarking-llms}), knowledge retrieval (\citealp{wei-etal-2025-teach}), time-aware translation (\citealp{chang-etal-2021-time}), and shared task (\citealp{wang-etal-2023-evahan2023}).

\section{The Expressiveness of the \textit{Kundoku} Mark System}

As introduced in the previous section, \textbf{Kaeriten} marks dictate the reading order of Chinese characters in Japanese translation. Since classical Chinese and Japanese have the syntactic divergence, it is crucial to investigate how many permutations of Chinese characters can be expressed by \textbf{Kaeriten} marks.~\citet{shimano-2012-permutation} approached this question from the perspective of combinatorics. In this section, we abstract the reading process into stack operations and address the same problem from a computational perspective. Based on this abstraction, we implement a pushdown automaton (PDA) that decodes Chinese characters with annotations.

\subsection{Reading via a Stack Data Structure}

\begin{CJK}{UTF8}{ipxm}
\begin{figure}[t]
    \centering
    \begin{tikzpicture}[x = 0.75pt, y = 0.75pt, yscale = -1, xscale = 1]
        \fill [fill = cyan!40] (-25, 10) rectangle (0, 28);
        \fill [fill = green!40] (0, 10) rectangle (25, 28);
        \fill [fill = orange!40] (25, 10) rectangle (40, 28);
        \fill [fill = cyan!40] (154, 10) rectangle (167, 28);
        \fill [fill = green!40] (141, 10) rectangle (154, 28);
        \fill [fill = orange!40] (127, 10) rectangle (141, 28);
        \draw[->,thick,rounded corners = 2pt, black] (-12.5, 28) -- (-10, 35) -- (8,35)-- (10.5, 28);
        \draw[->,thick,rounded corners = 2pt, black] (14.5, 28) -- (17, 35) -- (30,35)-- (32.5, 28);
        \node [text = black, align = left, font = \normalsize] at (0, 0) {Input Characters};
        \node [text = black, align = left, font = \normalsize] at (140, -1.6) {Ordered Characters};
        \node [text = black, align = left, font = \normalsize] at (0, 20) {A B\textsubscript{レ} C\textsubscript{レ} D};
        \node [text = black, align = left, font = \normalsize] at (140, 18.4) {A D C B};
        \node [text = black, align = left, font = \normalsize] at (75, 50) {Operations: Push(B), Push(C), Pop(C), Pop(B)};
    \end{tikzpicture}

    \vspace{0.5em}

    \begin{tikzpicture}[x = 0.75pt, y = 0.75pt, yscale = -1, xscale = 1]
        \fill [fill = cyan!40] (-55, 10) rectangle (-32, 28);
        \fill [fill = red!40] (-32, 10) rectangle (-7, 28);
        \fill [fill = green!40] (7, 10) rectangle (32, 28);
        \fill [fill = orange!40] (32, 10) rectangle (55, 28);
        \fill [fill = cyan!40] (160, 10) rectangle (173, 28);
        \fill [fill = red!40] (134, 10) rectangle (147, 28);
        \fill [fill = green!40] (120, 10) rectangle (134, 28);
        \fill [fill = orange!40] (147, 10) rectangle (160, 28);
        \draw[->,thick,rounded corners = 2pt, black] (-43.5, 28) -- (-41, 40) -- (41,40)-- (43.5, 28);
        \draw[->,thick,rounded corners = 2pt, black] (-19.5, 28) -- (-17, 35) -- (19.5,35)-- (22, 28);
        \node [text = black, align = left, font = \normalsize] at (0, 0) {Input Characters};
        \node [text = black, align = left, font = \normalsize] at (140, -1.6) {Ordered Characters};
        \node [text = black, align = left, font = \normalsize] at (0, 20) {A\textsubscript{二} B\textsubscript{下} C D\textsubscript{上} E\textsubscript{一}};
        \node [text = black, align = left, font = \normalsize] at (140, 18.4) {C D B E A};
        \node [text = black, align = left, font = \normalsize] at (75, 50) {Operations: Push(A), Push(B), Pop(B), Pop(A)};
    \end{tikzpicture}

    \vspace{0.5em}

    \begin{tikzpicture}[x = 0.75pt, y = 0.75pt, yscale = -1, xscale = 1]
        \fill [fill = cyan!40] (-44, 10) rectangle (5, 28);
        \fill [fill = green!40] (19, 10) rectangle (43, 28);
        \fill [fill = cyan!40] (140, 10) rectangle (167, 28);
        \fill [fill = green!40] (126, 10) rectangle (140, 28);
        \draw[->,thick,rounded corners = 2pt, black] (-19.5, 28) -- (-17, 35) -- (31,35)-- (33.5, 28);
        \node [text = black, align = left, font = \normalsize] at (0, 0) {Input Characters};
        \node [text = black, align = left, font = \normalsize] at (140, -1.6) {Ordered Characters};
        \node [text = black, align = left, font = \normalsize] at (0, 20) {A\textsubscript{二}—B C D\textsubscript{一}};
        \node [text = black, align = left, font = \normalsize] at (140, 18.4) {C D A B};
        \node [text = black, align = left, font = \normalsize] at (75, 50) {Operations: Push(AB), Pop(AB)};
    \end{tikzpicture}
    \caption{Examples of \textbf{Kaeriten}. Sentences on the left are characters with marks. Sentences on the right are characters in the correct order. Black arrows represent characters being read after the target character. Stack operations are listed under each example.}
    \label{fig2}
\end{figure}
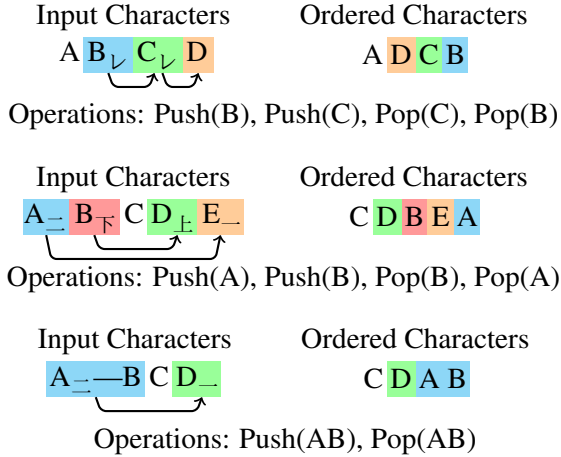
\end{CJK}

Consider a sequence of Chinese characters annotated with \textbf{Kaeriten}. For those characters without annotations, their relative position remains unchanged in the final reordered sequence. Therefore, they require no stack operations. For characters with \textbf{Kaeriten} marks, their positions are altered. Based on the functions, \textbf{Kaeriten} marks can be categorized into three types, as shown below:

\begin{CJK}{UTF8}{ipxm}
\paragraph{レ mark.} This mark indicates that the character should appear immediately after its successor in the reordered sequence. As shown in the top example of Figure~\ref{fig2}, character B bears a レ mark and is followed by character C. Thus, in the target sequence, B appears immediately after C. In terms of stack operations, we push a character onto the stack, if it holds a レ mark. This character should be popped from the stack, when its following character appears at the end of currently decoded sequence.

\paragraph{一二 like mark.} This category encompasses ordinal marks such as 一二, 上下, and 甲乙. This type of marks explicitly dictate the reading order. In the middle example of Figure~\ref{fig2}, character A appears immediately after character E, while character B tightly succeeds character D. In stack operations, we push a character annotated with 二, 下 or 乙 onto the stack, when we encounter it. The character is popped from the stack, when its corresponding character with 一, 上 or 甲 has been read and appended to the end of currently decoded sequence. It is worth noting that this type of mark enforces a hierarchy structure. Characters with 上下 must be nested within characters with 一二, as shown in Figure~\ref{fig2}. This hierarchy ensures the Last-in First-out (LIFO) principle of stack is preserved during the annotation process.

\paragraph{— mark.} This mark functions as a hyphen in English. It serves to connect characters into a single unit. Consequently, other \textbf{Kaeriten} marks operate on this unit as a whole instead of an individual character. As shown in the bottom example of Figure~\ref{fig2}, characters A and B are moved, pushed, and popped as a single unit.
\end{CJK}

\vspace{0.5em}

Except for \begin{CJK}{UTF8}{ipxm}—\end{CJK} mark, the other \textbf{Kaeriten} marks operate at the individual character level and are directly related to stack operations. Thus, the number of character sequences that can be expressed by the \textbf{Kaeriten} mark system is theoretically equivalent to the number of sequences that can be sorted by a stack.~\citet{knuth-1997-art} discussed the enumeration of such stack sortable permutations. For a sequence of $n$ characters (yielding $n!$ permutations), by Knuth's theorem, the number of sequences that can be expressed by \begin{CJK}{UTF8}{ipxm}レ mark and 一二 like mark\end{CJK} is
\begin{equation*}
    \textstyle
    a_{n} = \tbinom{2n}{n} - \tbinom{2n}{n - 1} = \frac{1}{n + 1}\tbinom{2n}{n}
\end{equation*}

When we incorporate \begin{CJK}{UTF8}{ipxm}—\end{CJK} mark, the analysis becomes more complicated, since each element in the stack can represent a sequence of characters.~\citet{Atkinson-2002-permutation} referred to this data structure as \textit{a stack of queues}. They derived the generating function of this data structure, which is
\begin{equation*}
    \textstyle
    \frac{1 - 3x + x^2 - \sqrt{1 - 6x + 7x^2 - 2x^3 + x^4}}{2x}
\end{equation*}
\citet{kruchinin-2013-compositaproperties} provided the methodology for deriving the closed-form formula from this generating function. By their theory, the number of sequences among $n!$ permutations of $n$ characters that can be expressed by \textbf{Kaeriten} is 
\begin{equation*}
\resizebox{\columnwidth}{!}{$
    a_{n} = \sum_{m = 1}^{n + 1} \frac{(-1)^{n - m + 1}}{m} \tbinom{m}{n - m + 1} \left(\sum_{i = 0}^{m - 1}\tbinom{m}{i}\tbinom{2m - i - 2}{m - 1}\right)
$}
\end{equation*}
This formula represents the theoretical upper bound of the expressiveness of the \textbf{Kaeriten} mark system.\footnote{For more information about the generating function and the formula, see~\url{https://oeis.org/A078482}.}

{\setlength{\tabcolsep}{3pt}
    \begin{table}[h]
        \centering
        \begin{tabular}{cccc}
            \hline
            \#Chars & \#Perms & \#Perms by \textbf{Kaeriten} & Pct \\
            \hline
            1 & 1 & 1 & 100\\
            2 & 2 & 2 & 100\\
            3 & 6 & 6 & 100\\
            4 & 24 & 20 & 83.33\\
            5 & 120 & 70 & 58.33\\
            6 & 720 & 254 & 35.28 \\
            7 & 5040 & 948 & 18.81 \\
            8 & 40320 & 3618 & 9.97 \\
            \hline
        \end{tabular}
    \caption{Considering $n$ Chinese characters, the number and percentage of permutations of characters that can be expressed by \textbf{Kaeriten} marks.}
    \label{table1}
    \end{table}
}

Table~\ref{table1} describes the number of permutations of $n$ Chinese characters that can be expressed by \textbf{Kaeriten} marks. As the sequence length grows, the total number of permutations increases, while the proportion of acceptable permutations diminishes. This phenomenon suggests that natural language syntax imposes strong structural constraints, in spite of large-scale potential character permutations. When people do translation, the constraint by \textbf{Kaeriten} marks reduces the reasoning space. The impact of the constraint becomes more pronounced as sentence length increases.

\subsection{Automaton and Transducer}

Since the reading of \begin{CJK}{UTF8}{ipxm}レ mark and 一二 like mark\end{CJK} can be described as stack operations, we define a Pushdown Automaton (PDA) and transduction operations to read a list of Chinese characters and output them in the target Japanese order. \begin{CJK}{UTF8}{ipxm}— mark\end{CJK} can be easily incorporated into this PDA by regarding characters connected by it as single units in the input alphabet. This PDA reads character-mark pairs and only accepts sentences with valid annotation. 

\begin{CJK}{UTF8}{ipxm}
\begin{center}
\textbf{PDA for Kaeriten reading}

$L = \{\text{valid}\ (cm)^n \mid n \ge 1, c \in C, m \in M\}$

where $C = \{\mathrm{Chinese\ characters}\}$

and $M = \{E, \text{レ}, O_{i,j}, \text{—}\}$ \footnote{$E$ represents the case that there is no \textbf{Kaeriten} mark following the character. $O_{i,j}$ represents 一二 like mark with hierarchy number $i$ and ordinal number $j$. For instance, 一 is in the first layer of the hierarchy, and it is the first in its group. It is represented as $O_{1,1}$. Similarly, 二 is represented as $O_{1,2}$.}
\end{center}

\begin{itemize}
    \setlength{\itemsep}{0.5em}
    \item \textbf{States}: $Q = \{q_0, q_1, q_2, q_3, q_4\}$
    \item \textbf{Input Alphabet}: $\Sigma = C \cup M$
    \item \textbf{Stack Alphabet}: $\Gamma = \Sigma \cup \{Z_0\}$
    \item \textbf{Initial State}: $q_0$
    \item \textbf{Accepting States}: $F = \{q_4\}$
    \item \textbf{Initial Stack Symbol}: $Z_0$
\end{itemize}

\vspace{0.2em}

\textbf{Transition Function ($\delta$) and Transduction:}

\begin{table}[h]
    \centering
    \renewcommand{\arraystretch}{1.2}
    \begin{tabularx}{\columnwidth}{l >{\raggedright\arraybackslash}X @{\enspace$\rightarrow$\enspace} r}
        1) & $\delta(q_0, c, \sigma) = \{(q_0, c \sigma)\},\ \sigma \neq O_{i,1}$ & $\epsilon$\\
        2) & $\delta(q_0, c, O_{i,1}) = \{(q_3, O_{i,1})\}$ & $\epsilon$\\
        3) & $\delta(q_0, \text{レ}, \sigma) = \{(q_0, \text{レ} \sigma)\}$ & $\epsilon$\\
        4) & $\delta(q_0, E, c \sigma) = \{(q_1, \sigma)\}$ & $c$\\
        5) & $\delta(q_0, O_{i,j}, \sigma) = \{(q_0, O_{i,j} \sigma)\},\ j>1$ & $\epsilon$\\
        6) & $\delta(q_0, O_{i,1}, \sigma) =$\par $\{(q_0, O_{i,1} \sigma), (q_2, O_{i,1} \sigma)\}$ & $\epsilon$\\
        7) & $\delta(q_1, \epsilon, \text{レ} c \sigma) = \{(q_1, \sigma)\}$ & $c$\\
        8) & $\delta(q_1, \epsilon, Z_0) = \{(q_0, Z_0)\}$ & $\epsilon$\\
        9) & $\delta(q_1, \epsilon, \text{レ} O_{i,1}) = \{(q_2, O_{i,1})\}$ & $\epsilon$\\
        10) & $\delta(q_1, \epsilon, O_{i,j}) = \{(q_0, O_{i,j})\}$ & $\epsilon$\\
        11) & $\delta(q_2, \epsilon, O_{i,j} c O_{i,j + 1}) = \{(q_2, O_{i,j + 1})\}$ & $c$\\
        12) & $\delta(q_2, \epsilon, O_{i,j} c O_{m,n}) =$ \par $\{(q_0, O_{m,n})\},\ i \neq m$ & $c$\\
        13) & $\delta(q_2, \epsilon, O_{i,j} c Z_0) = \{(q_0, Z_0)\}$ & $c$\\
        14) & $\delta(q_2, \epsilon, O_{i,j} c \text{レ}) = \{(q_1, \text{レ})\}$ & $c$\\
        15) & $\delta(q_0, \epsilon, Z_0) = \{(q_4, Z_0)\}$ & $\epsilon$\\
        \multicolumn{2}{l}{note: $\sigma$ represents any element in $\Gamma$}
    \end{tabularx}
\end{table}
\end{CJK}

The PDA defined above serves as a mechanism to check the validity of \textbf{Kaeriten} annotations. The derived transducer facilitates the reordering of Chinese characters. By appending \textbf{Okurigana} marks to each reordered character, we obtain the final Japanese translation of the given classical Chinese sentence.

For a better understanding of \textbf{Kaeriten} marks and the execution of the PDA, we provide an illustrative example in Appendix~\ref{appendix3}.

\section{Dataset Construction}

The research on \textit{Kundoku} translation system faces a low-resource challenge. There is only one open source dataset created by~\citet{wang-etal-2023-kanbun}. This dataset contains approximately 3,400 sentences, restricted to the genre of classical Chinese poems from the 7\textsuperscript{th} century A.D. to the 10\textsuperscript{th} century A.D. Their dataset also lacks annotations. This dataset is insufficient for research on the \textit{Kundoku} system.

This section introduces how we constructed a new dataset from the largest online accessible website\footnote{\url{https://kanbun.info/}} about classical Chinese and Japanese translations. The data within this website is organized as pairs consisting of a punctuated classical Chinese sentence and its corresponding Japanese translation. However, this dataset still lacks \textbf{Okurigana} and \textbf{Kaeriten} annotations. Therefore, we propose methods to automatically generate these marks.

\subsection{Mark Generation}

In the \textit{Kundoku} system, \textbf{Okurigana} are Japanese kanas appearing in conjunction with Chinese characters. For most characters, we simply assign the immediately following kanas as their \textbf{Okurigana}. However, some Chinese characters are rendered as kanas in Japanese translations. We need to restore these to their original Chinese characters before further processing. Since these characters typically function as interjections or conjunctions in sentences, we identified their positions via part of speech (POS) tagging and constructed a dictionary to map the kanas back to Chinese characters. Since currently available Japanese POS taggers are based on Modern Japanese, to obtain reliable POS tags for the classical Japanese sentences, we first utilized the GiNZA POS tagger~\citep{GiNZA} to generate raw POS tags. Then, we refined the result using GPT-4o.

In our constructed dataset, we provided a more fine-grained annotation scheme. We distinguished kanas playing grammatical roles from \textbf{Okurigana} and classified them as \textbf{particle}. Different from \textbf{Okurigana}, \textbf{particle} is less related with specific Chinese characters and serves to indicate case, tense and so on. This distinction was also achieved through POS tagging analysis. For kanas appearing within the boundary of Japanese words with tag \texttt{VERB}, \texttt{ADV}, \texttt{NOUN}, \texttt{ADJ}, \texttt{PRON} and \texttt{DET}, we regarded them as \textbf{Okurigana}.

To generate \textbf{Kaeriten} marks, we first aligned each Chinese character in the original classical Chinese sentence with its counterpart in the Japanese translation to determine the reading order. There exists two methods to generate \textbf{Kaeriten} marks: building the inverse of the PDA described above, or employing a rule-based approach based on characters' relative positions. As illustrated in Figure~\ref{fig2}, we assign \begin{CJK}{UTF8}{ipxm}レ mark\end{CJK} to consecutive characters if they are reversed in the Japanese sentence. We assign \begin{CJK}{UTF8}{ipxm}一二 like mark\end{CJK} to non-consecutive characters if they appear together in the Japanese sentence with inverted order. We use \begin{CJK}{UTF8}{ipxm}— mark\end{CJK} to connect characters when they are reordered as a single unit. For the sake of simplicity, we selected the latter method.

\subsection{Dataset Statistics and Validation}

We collected 9,292 sentences (95,066 characters) of classical Chinese to form the dataset. The dataset covers genres including history, philosophy, military strategy and poetry, across different time periods. Regarding sentence length distribution, 6,099 sentences contain fewer than 10 characters; 2,667 range between 10 and 20; 388 range between 20 and 30; and 138 exceed 30 characters. Our dataset exhibits a better genre coverage and a balanced length distribution, making it highly valuable for the research on the \textit{Kundoku} system.

During the generation of \textbf{Kaeriten} marks, we deployed the PDA to check the quality of annotation. Almost all marks generated are accepted by the PDA, except for a few sentences, which contain errors in the original Japanese translations. We manually corrected these sentences and reannotated them. The high acceptance rate demonstrates the validity of our \textbf{Kaeriten} marks generation method. This result also underscores the nature of the historical annotation system and the syntactic alignment between classical Chinese and Japanese.

For generated \textbf{Okurigana} and \textbf{particle} marks, we also manually checked their quality and corrected mistakes. In our dataset, there are 42,473 characters with \textbf{Okurigana} or \textbf{particle} marks. Among them, we corrected 11,834 characters. The acceptance rate of LLM's annotation is 72.14\%. For \textbf{Okurigana} and \textbf{particle} generation, LLM-based approach is feasible, but still needs human intervention.

\section{Modeling and Ablation Study}

Since \textit{Kundoku} marks are assigned to each individual Chinese character, we formulate the annotation process as sequence tagging tasks and integrate them with modern NLP paradigms. After acquiring all marks, we employ the transducer and construct Japanese translations. There exist BERT/RoBERTa based language models pretrained on classical Chinese, such as sikubert~\citep{li-etal-2022-first}, bert-ancient-chinese~\citep{bert-ancient-chinese} and roberta-classical-chinese-base-char~\citep{roberta-classical-chinese-char}. These models with deep encoder architecture and character level tokenization are naturally suitable for the sequence labeling tasks.

Linguistic knowledge of Classical Chinese is dispensable for human annotators. Consequently, exploring the impact of injecting extra Chinese knowledge is an interesting research topic.

In this section, we fine-tune several pretrained language models on the dataset constructed above. We adopt a multi-task learning strategy when annotating distinct types of \textit{Kundoku} marks. Furthermore, we conduct an ablation study to quantify the effect of introducing extra linguistic knowledge via auxiliary learning tasks. We evaluate performance of models with machine translation metrics, sequence matching metrics and the PDA pass rate.

\subsection{Model Architecture}

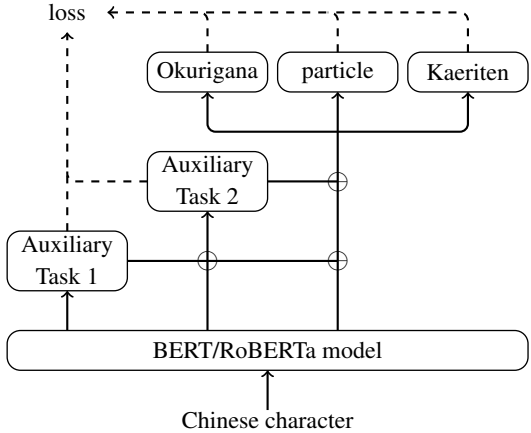
\begin{figure}[t]
    \centering
    \begin{tikzpicture}[x = 0.75pt, y = 0.75pt, yscale = 1, xscale = 1]
        \draw[draw = black, line width = 0.5, rounded corners=5pt] (0, 0) rectangle (260, 20);
        \draw[draw = black, line width = 0.5, rounded corners=5pt] (0, 40) rectangle (60, 70);
        \draw[draw = black, line width = 0.5, rounded corners=5pt] (70, 80) rectangle (130, 110);
        \draw[draw = black, line width = 0.5, rounded corners=5pt] (70, 140) rectangle (130, 160);
        \draw[draw = black, line width = 0.5, rounded corners=5pt] (135, 140) rectangle (195, 160);
        \draw[draw = black, line width = 0.5, rounded corners=5pt] (200, 140) rectangle (260, 160);
    
        \node [text = black, align = center, font = \small] at (130, 10) {BERT/RoBERTa model};
        \node [text = black, align = center, font = \small] at (30, 62.5) {Auxiliary};
        \node [text = black, align = center, font = \small] at (30, 47.5) {Task 1};
        \node [text = black, align = center, font = \small] at (100, 102.5) {Auxiliary};
        \node [text = black, align = center, font = \small] at (100, 87.5) {Task 2};
        \node [text = black, align = center, font = \small] at (100, 150) {Okurigana};
        \node [text = black, align = center, font = \small] at (165, 150) {particle};
        \node [text = black, align = center, font = \small] at (230, 150) {Kaeriten};
        \node [text = black, align = center, font = \small] at (30, 180) {loss};
        \node [text = black, align = center, font = \small] at (130, -25) {Chinese character};

        \draw[->, thick, black] (30, 20) -- (30, 40);
        \draw[->, dashed, thick, black] (30, 70) -- (30, 170);
        \draw[->, thick, black] (100, 20) -- (100, 80);
        \draw[->, thick, black] (165, 20) -- (165, 140);
        \draw[->, thick, black, rounded corners = 3pt] (165, 120) -- (100, 120) -- (100, 140);
        \draw[->, thick, black, rounded corners = 3pt] (165, 120) -- (230, 120) -- (230, 140);
        \draw[-, dashed, thick, black, rounded corners = 3pt] (100, 160) -- (100, 180) -- (95, 180);
        \draw[->, dashed, thick, black, rounded corners = 3pt] (165, 160) -- (165, 180) -- (50, 180);
        \draw[-, dashed, thick, black, rounded corners = 3pt] (230, 160) -- (230, 180) -- (160, 180);
        \draw[-, dashed, thick, black, rounded corners = 3pt] (70, 95) -- (30, 95) -- (30, 100);
        \draw[-, thick, black, rounded corners = 3pt] (60, 55) -- (165, 55);
        \draw[-, thick, black, rounded corners = 3pt] (130, 95) -- (165, 95);
        \draw[->, thick, black, rounded corners = 3pt] (130, -20) -- (130, 0);

        \node [text = black, align = center, font = \bfseries] at (100, 55) {$\oplus$};
        \node [text = black, align = center, font = \bfseries] at (165, 55) {$\oplus$};
        \node [text = black, align = center, font = \bfseries] at (165, 95) {$\oplus$};
    \end{tikzpicture}
    \caption{An example of our multitask learning model's structure. Solid lines represent the flow of embeddings and logits. Dashed lines represent the flow of loss.}
    \label{fig3}
\end{figure}

We adopted a joint learning strategy to predict all mark types simultaneously, and ensured that different classifiers share the same low-level representations so that they may benefit from each other. For auxiliary Chinese NLP tasks, we constructed a multi-step learning architecture~\citep{zhang-etal-2023-survey}, analogous to the multi-task learning framework by~\citet{hashimoto-etal-2017-joint}.

Figure~\ref{fig3} depicts the architecture of an example model with two auxiliary tasks. The output embeddings from the BERT/RoBERTa model are first passed to the classifier of Auxiliary Task 1 to compute the logits. Then, we concatenate the original output embeddings with these logits to form the input for Auxiliary Task 2. Finally, we concatenate the output embeddings with the logits from both Auxiliary Task 1 and Auxiliary Task 2 to form the final input for the three main classifiers. The total model loss is calculated as the weighted sum of the losses derived from all auxiliary tasks and main tasks, as shown below.
\begin{equation*}
    \begin{split}
        L_{model} =\ &w_{a}(L_{\text{aux task 1}} + L_{\text{aux task 2}}) +\\&w_{m}(L_{\text{Okurigana}} + L_{\text{particle}} + L_{\text{Kaeriten}})
    \end{split}
\end{equation*}

\subsection{Main Tasks and Auxiliary Tasks}

The primary labeling tasks consist of \textbf{Okurigana}, \textbf{particle} and \textbf{Kaeriten} prediction. We adopted marks appearing in our dataset as the label space of these classifiers. In our experiments, \textbf{Okurigana}, \textbf{particle} and \textbf{Kaeriten} have 434, 476 and 20 labels respectively.

In addition to the three main tasks, we adopted four classical Chinese sequence labeling tasks as auxiliary tasks. They are listed below:

\paragraph{Word Segmentation} Identifies word boundaries. We use the simplest B-I tag system in experiments.

\paragraph{Part Of Speech Tagging} Identifies grammatical roles. We use the Universal Dependencies POS tag system. In experiments, there are 14 labels.

\paragraph{Dependency Arc Labeling} Indicates the relative position of a word in the sentence's dependency tree. We utilize the tag system proposed by~\citet{gomez-rodriguez-etal-2023-4}. There are 115 labels in our dataset.

\paragraph{Dependency Type Labeling} Mark the type of dependency arc for each word. 44 labels are used in experiments.

\vspace{0.5em}

All auxiliary labels were generated based on the parse results obtained via HanLP\footnote {\url{https://github.com/hankcs/HanLP}}\citep{he-choi-2021-stem}, which offers classical Chinese dependency parsing. We introduced a special \texttt{continue} label across POS, dependency arc, and dependency type tasks to designate characters that are not the initial of a word. When computing loss, we took \texttt{continue} into consideration. Since in classical Chinese, most characters work as a word independently, including \texttt{continue} does not hinder training, and even forces models to learn to distinguish word boundaries.

All auxiliary tasks are organized according to the following hierarchy: word segmentation < POS tagging < dependency arc labeling < dependency type labeling. Regardless of the number of auxiliary tasks employed, this relative priority is always maintained.

\subsection{Evaluation Metrics}

We employ the same metrics used by~\citet{wang-etal-2023-kanbun} to ensure comparability of experimental results. These metrics are classified into machine translation metrics including \textbf{BLEU}~\citep{papineni-etal-2002-bleu}, \textbf{RIBES}~\citep{isozaki-etal-2010-automatic}, \textbf{ROUGE-L}~\citep{lin-2004-rouge}, \textbf{BERTScore}~\citep{bertscore}, and character reordering/matching metrics including \textbf{Kendall’s Tau ($\tau$)} and  \textbf{Perfect Match Ratio (PMR)}. All metrics are computed at the character level.

In addition to these metrics, we incorporate \textbf{chrF (character-level F-score)}~\citep{popovic-2015-chrf}, \textbf{TER (Translation Edit Rate)}~\citep{snover-etal-2006-study}, and the \textbf{pass rate of PDA} into the evaluation. The former two metrics assess the quality of Japanese translation, while the latter one evaluates language models' understanding of \textbf{Kaeriten} marks.

\subsection{Experimental Setting and Result}

\begin{table*}[t]
    \centering
    \resizebox{\textwidth}{!}{
    \begin{tabular}{lccccccccc}
        \hline
        Tasks & BLEU & chrF & BERTScore & ROUGE-L & RIBES & Kendall's $\tau$ & TER & PMR & Pass Rate \\
        \hline
        \textit{Kundoku} marks & 64.76 & 59.76 & 94.57 & 85.75 & 59.01 & 19.87 & 97.00 & 94.39 & 94.09\\
        +seg & 62.70 & 57.10 & 94.27 & 84.89 & 56.31 & 20.83 & 97.19 & 94.50 & 94.62\\
        +pos & 65.41 & 58.81 & 94.56 & 85.48 & 56.56 & 19.93 & 96.88 & 94.63 & \textbf{95.38}\\
        +arc & 64.99 & 59.69 & 94.51 & 85.52 & 57.96 & 19.93 & 97.27 & 94.84 & 92.37\\
        +type & 62.33 & 56.05 & 94.07 & 84.38 & 53.83 & 21.41 & \textbf{97.37} & \textbf{95.03} & 92.37\\
        +seg+pos & 65.24 & 59.71 & 94.54 & 85.58 & 58.01 & 19.86 & 96.80 & 94.51 & 92.69\\
        +seg+arc & 60.44 & 55.40 & 93.96 & 84.05 & 54.22 & 21.69 & 96.84 & 94.60 & 94.07\\
        +seg+type & \textbf{66.07} & \textbf{61.25} & \textbf{94.72} & \textbf{86.26} & \textbf{60.20} & \textbf{18.79} & 96.95 & 94.31 & 93.33\\
        +pos+arc & 63.06 & 58.40 & 94.22 & 84.85 & 57.65 & 20.57 & 96.96 & 94.42 & 93.23\\
        +pos+type & 61.90 & 58.29 & 94.22 & 85.20 & 59.25 & 20.62 & 97.07 & 94.49 & 94.84\\
        +arc+type & 63.03 & 59.09 & 94.31 & 85.17 & 58.75 & 20.32 & 96.81 & 94.17 & 94.41\\
        +seg+arc+type & 60.66 & 56.34 & 94.07 & 84.39 & 56.08 & 21.41 & 96.73 & 94.12 & 94.19\\
        +seg+pos+type & 62.22 & 57.44 & 94.26 & 84.88 & 56.41 & 20.71 & 96.78 & 94.01 & 94.19\\
        +seg+pos+arc & 61.17 & 56.17 & 94.16 & 84.51 & 55.06 & 21.41 & 96.66 & 94.02 & 93.98\\
        +pos+arc+type & 62.03 & 57.00 & 94.14 & 84.53 & 55.65 & 21.03 & 96.71 & 94.34 & 93.01\\
        +all & 62.14 & 57.09 & 94.13 & 84.69 & 56.9 & 21.19 & 96.49 & 93.93 & 92.15\\
        \hline
    \end{tabular}
    }
    \caption{Experimental result on roberta-classical-chinese-base-char.}
    \label{table2}
\end{table*}

The generated dataset was randomly shuffled and split into training, validation, and test sets with a ratio of 8:1:1. Experiments were conducted on sikubert, bert-ancient-chinese, roberta-classical-chinese-base-char and bert-base-japanese-char\footnote{\url{https://huggingface.co/tohoku-nlp/bert-base-japanese-char}}. For each pretrained language model, we trained a classifier to predict \textit{Kundoku} marks with 0, 1, 2, 3, 4 auxiliary classical Chinese NLP tasks incorporated. This resulted in a total of 16 different model configurations for each base language model. To ensure the main tasks remain the primary training objective, we adjusted the loss weights. We assigned main/auxiliary task weights of 8:2, 7:3, and 6:4 for configurations with one, two, and three or more auxiliary tasks, respectively. We adopted the early stopping strategy during training. For more details about the training setting, see Appendix~\ref{appendix1}.

Table~\ref{table2} presents the evaluation result for roberta-classical-chinese-base-char. Evaluation results for other models are available in Appendix~\ref{appendix5}. Overall, we observe that models based on pretrained classical Chinese models significantly outperform those models pretrained on Japanese. It emphasizes the significance of domain specific language knowledge acquired from pretraining. By comparing performance of models with auxiliary tasks configurations, we observe that Chinese NLP auxiliary tasks generally yield positive effects on the main tasks. Models based on sikubert, roberta-classical-chinese-base-char and bert-base-japanese-char achieve their best performance across many metrics with the inclusion of two auxiliary tasks. However, integrating additional auxiliary tasks leads to a decline in performance. We hypothesize that with more tasks introduced, the learning of main tasks is diluted. The model optimization becomes biased towards auxiliary tasks. Furthermore, with more auxiliary tasks included, the architecture of models gets deeper as well. This increased depth may impede optimization and loss propagation. Finally, the training data of auxiliary tasks was generated by dependency parsers. The parsers may have hallucination and introduce noise into the training data. This may lead to performance degradation.

\section{Model Comparison}

In this section, we compare our proposed approach against the translation pipeline proposed by~\citet{wang-etal-2023-kanbun}. Their models were evaluated on their classical Chinese poem dataset. To ensure the comparability, we applied identical data processing procedures to their dataset and trained a new model. Based on the ablation study results in the previous section, we used word segmentation and dependency type labeling as auxiliary tasks, since this combination of tasks reached the best machine translation performance. We trained this new model on roberta-classical-chinese-base-char and maintained the same experimental setting.

Table~\ref{table3} presents the evaluation result of our model alongside the best models built by~\citet{wang-etal-2023-kanbun}. Since they employed distinct models for characters reordering and machine translation, we report the best translation scores from mT5-large and best character ordering scores from roberta-classical-chinese-char. Overall, our approach achieves results comparable to the best models constructed by them. Notably, our model performs better in character ordering and has higher \textbf{perfect matching ratio (PMR)}. Consequently, our model achieves superior scores on the translation metrics, such as \textbf{RIBES}, which is sensitive to character order. In contrast to the two-stage pipeline, our approach has a smaller number of parameters and also provides character level annotations.

\begin{table*}[t]
    \centering
    \resizebox{\textwidth}{!}{
    \begin{tabular}{lccccccccc}
        \hline
        Models & BLEU & chrF & BERTScore & ROUGE-L & RIBES & TER & Kendall's $\tau$ & PMR & Pass Rate \\
        \hline
        our approach & 55.51 & 52.12 & 92.58 & 80.11 & 63.67 & 27.13 & 92.23 & 90.00 & 95.41 \\
        Wang et al. & 51.40 & - & 93.40 & 74.70 & 58.30 & - & 94.40 & 78.30 & - \\
        \hline
    \end{tabular}
    }
    \caption{Evaluation result on the dataset created by~\citet{wang-etal-2023-kanbun}. We trained a \textbf{new} model with POS tagging and dependency type labeling as auxiliary tasks and roberta-classical-chinese-base-char as the base model on their dataset.}
    \label{table3}
\end{table*}

\begin{table*}[t]
    \centering
    \resizebox{\textwidth}{!}{
    \begin{tabular}{clccccccccc}
        \hline
        & Models & BLEU & chrF & BERTScore & ROUGE-L & RIBES & TER & Kendall's $\tau$ & PMR & Pass Rate \\
        \hline
        & our approach & 62.50 & 57.93 & 94.38 & 84.96 & 55.72 & 21.07 & \textbf{97.35} & \textbf{94.91} & 92.00 \\
        \hline
        \multirow{4}{*}{\rotatebox{90}{zero-shot}} & DeepSeek-V3.2 & 61.96 & 57.60 & 93.73 & 82.60 & 53.52 & 23.46 & 82.49 & 62.28 & 91.00 \\
        & Gemini-3-pro-preview & 69.09 & 62.57 & 95.09 & 86.45 & 58.81 & 20.15 & 94.96 & 90.49 & 97.00 \\
        & Gemini-3-flash-preview & 68.72 & 61.88 & 94.73 & 85.70 & 56.58 & 18.86 & 94.02 & 89.72 & 97.00 \\
        & GPT-5.2 & 59.70 & 53.38 & 93.41 & 80.13 & 49.41 & 27.89 & 75.19 & 49.86 & 86.00 \\
        \hline
        \multirow{4}{*}{\rotatebox{90}{few-shot}} & DeepSeek-V3.2 & 64.23 & 57.42 & 94.34 & 83.58 & 53.05 & 22.33 & 77.77 & 58.22 & 89.00 \\
        & Gemini-3-pro-preview & \textbf{73.23} & \textbf{67.03} & \textbf{95.75} & 88.13 & \textbf{61.24} & \textbf{15.94} & 95.42 & 91.01 & 96.00 \\
        & Gemini-3-flash-preview & 72.79 & 65.86 & 95.56 & \textbf{88.16} & 59.74 & 16.11 & 95.13 & 91.21 & \textbf{98.00} \\
        & GPT-5.2 & 62.87 & 55.96 & 94.17 & 82.52 & 49.63 & 23.34 & 72.73 & 48.96 & 87.00 \\
        \hline
    \end{tabular}
    }
    \caption{Performance on Randomly selected 50 sentences in our test set. Machine translation metrics scores are evaluated on generated Japanese sentences. Character ordering scores are evaluated on generated \textbf{Kaeriten} marks after the processing of PDA. The middle part corresponds to the result of zero-shot experiments, while the lower part shows the result of few-shot experiments.}
    \label{table4}
\end{table*}

\section{How LLMs Perform on These Tasks}

Modern large language models (LLMs) have demonstrated their remarkable capabilities in understanding natural languages. It is crucial to assess their performance on our annotation and translation tasks to investigate their strengths and weaknesses of understanding classical Chinese and Japanese. Due to budget constraints, we randomly selected 100 sentences from our test set. We evaluated several most up-to-date models, including DeepSeek-V3.2, Gemini-3-pro-preview, Gemini-3-flash-preview, and GPT-5.2 on these sentences. We employed both zero-shot and few-shot prompting strategies to investigate the impact of in-context learning. First, we instructed LLMs to directly generate Japanese translations of classical Chinese sentences and evaluated the results with machine translation metrics. Subsequently, we prompted these LLMs to assign each Chinese character a \textbf{Kaeriten} mark and input their responses to the PDA. We evaluated their annotations on character ordering metrics. To keep consistent on model construction, we used the same base model and auxiliary tasks as used in the previous section.

Table~\ref{table4} outlines the comparative evaluation results between the LLMs and our approach. All LLMs achieve high scores in machine translation, with few-shot examples significantly enhancing their performance. Gemini models achieve the highest scores on machine translation metrics, while our approach is comparable to some LLMs, such as GPT-5.2 and DeepSeek-V3.2. However, regarding annotation metrics and character ordering, our approach outperforms all LLMs.

The high machine translation scores of LLMs might come from their training data. Since we collected and constructed our dataset from publicly available online resources, these sentences might be included in the training data of LLMs. Few-shot examples likely provide a context that assists LLMs in locating the correct translations. For \textbf{Kaeriten} annotation, since there is not enough training data available for LLMs, they have to rely on their reasoning abilities to understand \textbf{Kaeriten} marks and character ordering rules. Our model's translation is derived directly from the annotation results. Due to the limited label space, our model's translation output is not as flexible as LLMs. However, in our model, the character ordering and \textbf{Kaeriten} annotation benefit from the explicit supervision along with auxiliary Chinese NLP tasks. Consequently, our model achieves higher scores on character ordering metrics.

At the end of experiments, we applied our \textbf{Kaeriten} annotation pipeline to Japanese translations directly generated by the few-shot Gemini-3-pro-preview. We obtained new \textbf{Kaeriten} marks and did the same evaluation process. In this setting, the \textbf{Kendall's $\tau$} score and the \textbf{PMR} score improved to 96.18 and 93.72 respectively. This result indicates that LLMs possess the implicit character ordering knowledge from Chinese to Japanese, but they sometimes do not explicitly express in the annotation. Our annotation method can serve as an effective supplement to LLMs for correctly generating marks.

\section{Conclusion}

Japanese people translate classical Chinese into Japanese via annotation. In this work, we formulate this annotation process within the modern NLP paradigm as sequence tagging tasks. \textbf{Kaeriten} plays a central role in the syntactic reconstruction of Japanese sentence. We demonstrate that the annotation and reading of \textbf{Kaeriten} marks can be abstracted as sorting a sequence with a stack. We derive the theoretical upper bound of the expressiveness of \textbf{Kaeriten} and construct a pushdown automaton to validate annotation quality and generate reordered characters. To alleviate the low-resource problem, we construct a new dataset with annotations from online open source data. During the construction, we validate the effectiveness of the PDA and highlight the tight syntactic relation between classical Chinese and Japanese. Furthermore, we develop multi-task learning models and observe the benefit of introducing auxiliary Chinese NLP tasks. To achieve the best performance, empirical results suggest that the number of auxiliary tasks should not be more than two. Finally, we evaluate different LLMs on these tasks. Our evaluation of LLMs reveals that LLMs generate very high-quality translations, but there is still room for improvement in annotating \textbf{Kaeriten} marks.

\section*{Limitations}

Despite introducing a substantial dataset, the size of training data remains quite limited. We are still facing a low-resource challenge. We anticipate that by digitizing more books annotated by ancient people, we could address the low-resource problem. 

Our dataset is semi-automatically constructed using LLMs. Since LLMs are not well trained on classical languages, they introduce mistakes in the generated annotation. We still need to make an effort to correct generated marks. This process needs well trained annotator and time-consuming.

Additionally, our current classification of \textit{Kundoku} marks is based on linguistic intuition. This classification is coarse and not optimized for practice. Having a fine-grained set of marks might constrain the sampling space more effectively and yield superior results.

\section*{Acknowledgments}
We appreciate the work of Web Kanbun-Taikei, the biggest online resource available for \textit{Kundoku} research. The contributors collect \textit{Kanbun} documents and provide standard translations. We also thank the ARR reviewers and ACs for their insightful comments for improving the paper.

\begin{CJK}{UTF8}{ipxm}
\bibliography{custom}
\end{CJK}

\clearpage

\appendix

\section{Hyperparameters}
\label{appendix1}

The follow table describes hyperparameters we used when finetuning pretrained language models.

\begin{table}[h]
    \centering
    \begin{tabular}{lc}
        \hline
         Hypermeter & Value \\
         \hline
         batch size & 64 \\
         optimizer & AdamW \\
         beta1 & 0.9 \\
         beta2 & 0.999 \\
         epsilon & 1e-8 \\
         weight decay & 0.01 \\
         learning rate & 5e-5 \\
         dropout & 0.3 \\
         epoches & 15 \\
         early stopping epoch & 1 \\
         \hline
    \end{tabular}
    \caption{Hyperparameters of finetuning pretrained language models}
    \label{table5}
\end{table}

\section{LLM Prompts}
\label{appendix2}

We use the following prompt to make LLMs revise the POS tags generated by GiNZA:

\begin{tcolorbox}[colback = gray!10, colframe = black, title = {Prompt}, breakable]
    \begin{CJK}{UTF8}{ipxm}
        あなたは日本語の専門家です。次は漢文、書き下し文及び書き下し文の品詞タグ付きです。これに基づき、書き下し文の品詞タグを訂正してください。GiNZA品詞体系を使ってください。動詞と名詞の語幹及び送り仮名をに注意してください。助詞、例えば「て」、助動詞、例えば「し」に注意してください。再読文字、例えば「未だ」「将に」「須らく」「若お」「蓋ぞ」、とそれらの送り仮名に注意してください。形式名詞に対して、「名詞-普通名詞-形式名詞」のタグを使ってください。

        ---------
        
        <Classical Chinese Sentence>
        
        <Japanese Translation>
    \end{CJK}
\end{tcolorbox}

\noindent We use the following zero-shot prompt and few-shot prompt to let LLMs generate Japanese translation and \textbf{Kaeriten} marks towards provided classical Chinese sentences:

\begin{tcolorbox}[colback = gray!10, colframe = black, title = {Zero-shot Prompt}, breakable]
    \begin{CJK}{UTF8}{ipxm}
        あなたは日本語と古典中国語の専門家です。次の漢文を日本語の書き下し文に翻訳してください。翻訳すると共に、一つ一つ漢字と句読点に付く返り点を書いてください。返り点を付く必要がない場合、empty string ""を使てください。また、複数の返り点が現れる場合、一つのstringに書いてください。例えば、"一"と"レ"を"一レ"に書きます。漢字の振り仮名を付けないでください。

        ---------

        次の漢文を読んで書き下し文と返り点を書いてください。

        <classical Chinese sentence>。この漢文には<the length of classical Chinese sentence>の漢字と句読点があります。それぞれの漢字と句読点に対して返り点を提出してください。
    \end{CJK}
\end{tcolorbox}

\begin{tcolorbox}[colback = gray!10, colframe = black, title = {Few-shot Prompt}, breakable]
    \begin{CJK}{UTF8}{ipxm}
        あなたは日本語と古典中国語の専門家です。次の漢文を日本語の書き下し文に翻訳してください。翻訳すると共に、一つ一つ漢字と句読点に付く返り点を書いてください。返り点を付ける必要がない場合、必ずempty string ""を使てください。また、複数の返り点が現れる場合、一つのstringに書いてください。例えば、"一"と"レ"を"一レ"に書きます。漢字の振り仮名を付けないでください。

        ---------

        例えば、漢文「如聞泣幽咽」の書き下し文は「泣きて幽咽するを聞くが如し」で、それぞれの漢字と句読点に対する返り点は"レ", "二", "", "", "一"です。

        漢文「子曰、富與貴、是人之所欲也。」の書き下し文は「子曰く、富と貴きとは、是れ人の欲する所なり。」で、それぞれの漢字と句読点に対する返り点は"", "", "", "", "", "", "", "", "", "", "レ", "", "", ""です。

        漢文「罰不遷列、欲民速覩爲不善之害也。」の書き下し文は「罰、列を遷さざるは、民の速やかに不善を為すの害を覩んことを欲すればなり。」で、それぞれの漢字と句読点に対する返り点は"", "レ", "レ", "", "", "二", "", "", "下", "乙", "", "甲", "", "上", "一", ""です。

        ---------

        次の漢文を読んで書き下し文と返り点を書いてください。

        <classical Chinese sentence>。この漢文には<the length of classical Chinese sentence>の漢字と句読点があります。それぞれの漢字と句読点に対して返り点を提出してください。
    \end{CJK}
\end{tcolorbox}

\section{An Example of the Kaeriten Marks and the Execution of the PDA}
\label{appendix3}

We use the following sentence as the example:
\begin{CJK}{UTF8}{ipxm}
\exdisplay
\begingl
    \gla  人 君 無 以 三 寶 借 人 //
    \glb \textit{r\'{e}n} \textit{j\={u}n} \textit{w\'{u}} \textit{y\v{i}} \textit{s\={a}n} \textit{b\v{a}o} \textit{j\`{i}e} \textit{r\'{e}n} //
    \glc people ruler don't take three treasures lend {other people} //
\endgl
\xe

\noindent The \textbf{Kaeriten} marks of this sentence is: 人君無\textsubscript{二}以\textsubscript{下}三宝\textsubscript{上}借\textsubscript{一レ}人

\noindent The process of this sentence is listed below:

\begin{enumerate}[noitemsep]
    \item reverse the adjacent characters with レ mark: 人君無\textsubscript{二}以\textsubscript{下}三宝\textsubscript{上}人借\textsubscript{一}
    \item put the character with 下 mark to the right of the character with 上 mark: 人君無\textsubscript{二}三宝以人借\textsubscript{一}
    \item shift the character with 二 mark to the right of the character with 一 mark: 人君三宝以人借無
\end{enumerate}

\noindent Finally, we get the Japanese reading order of these characters in the classical Chinese sentence. The gold standard Japanese translation of this sentence is 人君は三宝を以て人に借す無かれ. It confirms the validity of \textbf{Kaeriten} mark system.

\vspace{0.4em}

\noindent The following table~\ref{table6} demonstrates the correct execution path of the PDA. The input of this PDA is a sequence of characters with their \textbf{Kaeriten} marks. For those characters without \textbf{Kaeriten} annotated, we regard their \textbf{Kaeriten} as the special symbol $E$. In this case, the input sequence is ["人", $E$, "君", $E$, "無", "二", "以", "下", "三", $E$, "宝", "上", "借", "一", "レ", "人", $E$].

\begin{table*}[h!]
    \centering
    \resizebox{\textwidth}{!}{
    \begin{tabular}{ccccc}
        \hline
        State & Input & Stack & Transduced & Transition \\
        \hline
        $q_0$ & ["人", $E$, \dots] & [$Z_0$] & "" & (1) \\
        $q_0$ & [$E$, "君", \dots] & ["人", $Z_0$] & "" & (4) \\
        $q_1$ & ["君", $E$, \dots] & [$Z_0$] & "人" & (8) \\
        $q_0$ & ["君", $E$, \dots] & [$Z_0$] & "人" & (1) \\
        $q_0$ & [$E$, "無", \dots] & ["君", $Z_0$] & "人" & (4) \\
        $q_1$ & ["無", "二", \dots] & [$Z_0$] & "人君" & (8) \\
        $q_0$ & ["無", "二", \dots] & [$Z_0$] & "人君" & (1) \\
        $q_0$ & ["二", "以", \dots] & ["無", $Z_0$] & "人君" & (5) \\
        $q_0$ & ["以", "下", \dots] & ["二", "無", $Z_0$] & "人君" & (1) \\
        $q_0$ & ["下", "三", \dots] & ["以", "二", "無", $Z_0$] & "人君" & (5) \\
        $q_0$ & ["三", $E$, \dots] & ["下", "以", "二", "無", $Z_0$] & "人君" & (1) \\
        $q_0$ & [$E$, "宝", \dots] & ["三", "下", "以", "二", "無", $Z_0$] & "人君" & (4) \\
        $q_1$ & ["宝", "上", \dots] & ["下", "以", "二", "無", $Z_0$] & "人君三" & (10) \\
        $q_0$ & ["宝", "上", \dots] & ["下", "以", "二", "無", $Z_0$] & "人君三" & (1) \\
        $q_0$ & ["上", "借", \dots] & ["宝", "下", "以", "二", "無", $Z_0$] & "人君三" & (6) \\
        $q_2$ & ["借", "一", \dots] & ["上", "宝", "下", "以", "二", "無", $Z_0$] & "人君三" & (11) \\
        $q_2$ & ["借", "一", \dots] & ["下", "以", "二", "無", $Z_0$] & "人君三宝" & (12) \\
        $q_0$ & ["借", "一", \dots] & ["二", "無", $Z_0$] & "人君三宝以" & (1) \\
        $q_0$ & ["一", "レ", \dots] & ["借", "二", "無", $Z_0$] & "人君三宝以" & (6) \\
        $q_0$ & ["レ", 人", \dots] & ["一", "借", "二", "無", $Z_0$] & "人君三宝以" & (3) \\
        $q_0$ & ["人", $E$] & ["レ", "一", "借", "二", "無", $Z_0$] & "人君三宝以" & (1) \\
        $q_0$ & [$E$] & ["人", "レ", "一", "借", "二", "無", $Z_0$] & "人君三宝以" & (4) \\
        $q_1$ & [] & ["レ", "一", "借", "二", "無", $Z_0$] & "人君三宝以人" & (9) \\
        $q_2$ & [] & ["一", "借", "二", "無", $Z_0$] & "人君三宝以人" & (11) \\
        $q_2$ & [] & ["二", "無", $Z_0$] & "人君三宝以人借" & (13) \\
        $q_0$ & [] & [$Z_0$] & "人君三宝以人借無" & (15) \\
        $q_4$ & [] & [$Z_0$] & "人君三宝以人借無" & - \\
        \hline
    \end{tabular}
    }
    \caption{An example of the PDA execution.}
    \label{table6}
\end{table*}

\end{CJK}

\section{The Process of \textit{Saidoku} (Read Again) Characters}
\label{appendix4}

\begin{CJK}{UTF8}{ipxm}

\textbf{Saidoku} (read again) characters are a special type of words in the classical Chinese translation. They are first read as a regular word without \textbf{Kaeriten} marks, then read as the word with \textbf{Kaeriten} annotated. For example, the following sentence annotated with \textbf{Kaeriten} marks 鼎之軽重、未\textsubscript{レ}可\textsubscript{レ}問也。 has Japanese translation 鼎の軽重、未だ問う可からざるなり。 In this sentence, 未だざる are \textbf{Saidoku} characters. Its Chinese character is first read as a regular character. Therefore, the characters 未だ appear following the punctuation 、. The rest part is then read as a word with \textbf{Kaeriten} mark レ. ざる shows after Chinese character 可 in the translation. The following algorithm~\ref{algorithm1} describes how we process \textbf{Saidoku} (read again) characters. In the execution of our proposed PDA, we could simply enforce the immediate output of characters when it reads \textbf{Saidoku} characters.

\end{CJK}

\begin{algorithm*}[h!]
\caption{Reading \textbf{Saidoku} Characters}
\begin{algorithmic}[1] 
\Procedure{Reading Chinese Characters}{Input Sequence $I$, Output Sequence $O$, PDA $A$}
    \While{$A$ does not terminate}
        \If{$A$ accepts a new input character} \Comment{steps consume a new input character}
            \State $c \gets A[0]$ \Comment{get the input character}
            \State $A \gets A[1:]$
            \If{$c$ is a \textbf{Saidoku} character}
                \State $O.\text{append}(c[\text{regular\ part}])$
                \State Execute $A$ to the next step with $c[\text{Saidoku\ part}]$ and $O$
            \Else
                \State Execute $A$ to the next step with $c$ and $O$
            \EndIf
        \Else \Comment{steps do not consume a new input character}
            \State Execute $A$ to the next step with $O$
        \EndIf
    \EndWhile
    \State \textbf{return} $O$
\EndProcedure
\end{algorithmic}
\label{algorithm1}
\end{algorithm*}

\section{Ablation Experiment Results}
\label{appendix5}

The following three tables~\ref{table7},~\ref{table8} and~\ref{table9} show the experimental results of our models with 0, 1, 2, 3, 4 auxiliary tasks. We ran 16 models on the base model selected from sikubert, bert-ancient-chinese and bert-base-japanese-char.

\begin{table*}[h!]
    \centering
    \resizebox{\textwidth}{!}{
    \begin{tabular}{lccccccccc}
        \hline
        Tasks & BLEU & chrF & BERTScore & ROUGE-L & RIBES & TER & Kendall's $\tau$ & PMR & Pass Rate \\
        \hline
        \textit{Kundoku} marks & 62.99 & 57.49 & 94.33 & 84.91 & 56.98 & 21.01 & 96.56 & 93.78 & 93.76\\
        +seg & 64.40 & 58.33 & 94.48 & 85.34 & 57.13 & 20.51 & 96.82 & 94.53 & 92.37 \\
        +pos & 62.54 & 58.14 & 94.16 & 84.71 & 56.69 & 20.86 & 96.58 & 93.73 & 93.55 \\
        +arc & 65.16 & 60.63 & \textbf{94.70} & \textbf{86.26} & \textbf{60.82} & 19.38 & 96.76 & 94.22 & 92.58 \\
        +type & \textbf{65.68} & 60.54 & 94.68 & 86.01 & 59.80 & 19.62 & 96.95 & 94.20 & 93.87 \\
        +seg+pos & 63.59 & 57.77 & 94.27 & 85.07 & 56.49 & 20.90 & 96.42 & 94.20 & 93.01 \\
        +seg+arc & 63.49 & 57.59 & 94.33 & 85.02 & 55.92 & 20.97 & \textbf{97.01} & \textbf{94.55} & 91.94 \\
        +seg+type & 63.47 & 56.52 & 94.28 & 84.94 & 54.10 & 21.34 & 96.76 & 94.08 & 93.01 \\
        +pos+arc & 64.33 & 59.67 & 94.49 & 85.57 & 58.98 & 19.98 & 96.37 & 93.84 & 92.90 \\
        +pos+type & 65.09 & \textbf{60.96} & 94.65 & 86.11 & 60.52 & \textbf{19.29} & 96.49 & 94.44 & 91.94 \\
        +arc+type & 62.57 & 58.23 & 94.29 & 85.09 & 59.46 & 20.97 & 96.34 & 93.52 & 92.47 \\
        +seg+arc+type & 61.59 & 56.91 & 94.11 & 84.82 & 56.99 & 21.34 & 96.32 & 93.31 & 91.29 \\
        +seg+pos+type & 63.00 & 57.70 & 94.41 & 85.19 & 56.95 & 20.93 & 96.32 & 93.33 & \textbf{94.62} \\
        +seg+pos+arc & 62.09 & 57.82 & 94.23 & 85.18 & 58.48 & 20.94 & 96.56 & 94.07 & 90.65 \\
        +pos+arc+type & 61.55 & 57.13 & 94.15 & 84.98 & 57.59 & 21.07 & 96.70 & 94.22 & 93.44 \\
        +all & 61.84 & 57.54 & 94.31 & 85.15 & 58.54 & 20.94 & 96.54 & 94.31 & 91.83 \\
        \hline
    \end{tabular}
    }
    \caption{Experimental results on sikubert.}
    \label{table7}
\end{table*}

\begin{table*}[h!]
    \centering
    \resizebox{\textwidth}{!}{
    \begin{tabular}{lccccccccc}
        \hline
        Tasks & BLEU & chrF & BERTScore & ROUGE-L & RIBES & TER & Kendall's $\tau$ & PMR & Pass Rate \\
        \hline
        \textit{Kundoku} marks & 64.27 & 57.92 & 94.33 & 84.91 & 55.34 & 20.90 & 96.72 & 94.17 & \textbf{95.27} \\
        +seg & 63.76 & 59.30 & 94.35 & 85.51 & 58.95 & \textbf{19.13} & 96.90 & 94.70 & 90.65 \\
        +pos & 65.52 & 61.13 & \textbf{94.76} & \textbf{86.36} & 60.78 & 19.28 & \textbf{97.05} & \textbf{94.89} & 92.89 \\
        +arc & 63.80 & 58.15 & 94.32 & 85.08 & 55.81 & 20.58 & 96.38 & 93.88 & 93.44 \\
        +type & 64.38 & 59.94 & 94.46 & 85.56 & 59.62 & 20.01 & 96.92 & 94.08 & 93.76 \\
        +seg+pos & 64.46 & 58.21 & 94.36 & 85.46 & 56.26 & 20.57 & 96.78 & 94.79 & 91.07 \\
        +seg+arc & 65.21 & 60.08 & 94.65 & 85.96 & 59.44 & 19.67 & 96.93 & 94.75 & 93.01 \\
        +seg+type & \textbf{65.58} & 60.33 & 94.62 & 86.06 & 59.32 & 19.45 & 96.87 & 94.51 & 93.87 \\
        +pos+arc & 64.86 & 59.60 & 94.55 & 85.62 & 58.63 & 20.06 & 96.54 & 94.00 & 93.98 \\
        +pos+type & 63.45 & 58.43 & 94.35 & 85.22 & 58.79 & 20.67 & 96.62 & 94.19 & 94.09 \\
        +arc+type & 65.43 & \textbf{60.91} & 94.68 & 86.19 & \textbf{61.42} & \textbf{19.13} & 96.90 & 94.71 & 91.29 \\
        +seg+arc+type & 62.58 & 57.98 & 94.22 & 84.87 & 57.19 & 20.73 & 96.29 & 93.70 & 93.01 \\
        +seg+pos+type & 64.97 & 60.40 & 94.58 & 86.13 & 59.85 & 19.36 & 96.97 & 94.50 & 92.04 \\
        +seg+pos+arc & 62.61 & 58.73 & 94.32 & 85.17 & 58.80 & 20.49 & 96.56 & 93.95 & 92.37 \\
        +pos+arc+type & 61.94 & 57.39 & 94.17 & 84.85 & 56.93 & 21.12 & 96.69 & 93.94 & 91.61 \\
        +all & 62.46 & 57.16 & 94.14
        & 84.99 & 57.21 & 20.91 & 96.88 & 94.65 & 92.04 \\
        \hline
    \end{tabular}
    }
    \caption{Experimental results on bert-ancient-chinese.}
    \label{table8}
\end{table*}

\begin{table*}[h!]
    \centering
    \resizebox{\textwidth}{!}{
    \begin{tabular}{lccccccccc}
        \hline
        Tasks & BLEU & chrF & BERTScore & ROUGE-L & RIBES & TER & Kendall's $\tau$ & PMR & Pass Rate \\
        \hline
        \textit{Kundoku} marks & 52.34 & 47.50 & 92.52 & 79.82 & 46.75 & 27.53 & 93.71 & 88.75 & 87.85 \\
        +seg & 53.96 & 50.35 & 92.72 & 80.52 & 50.67 & 26.55 & 93.60 & 88.99 & 91.08 \\
        +pos & 54.58 & 49.93 & 92.78 & 80.69 & 49.69 & 26.27 & 93.47 & 88.92 & \textbf{91.72} \\
        +arc & 53.48 & 48.77 & 92.66 & 80.36 & 48.46 & 27.06 & 94.15 & 89.86 & 89.57 \\
        +type & 53.58 & 49.85 & 92.78 & 80.86 & 50.82 & 26.17 & 94.51 & 89.72 & 85.59 \\
        +seg+pos & 54.16 & 49.01 & 92.77 & 80.50 & 47.93 & 27.02 & 93.64 & 88.45 & 91.18 \\
        +seg+arc & 52.99 & 49.27 & 92.71 & 80.36 & 49.71 & 26.74 & 93.57 & 89.09 & 88.17 \\
        +seg+type & 53.93 & 49.79 & 92.75 & 80.76 & 49.15 & 26.49 & 93.90 & 89.43 & 90.54 \\
        +pos+arc & 53.96 & 50.62 & 92.76 & 80.93 & \textbf{52.17} & 26.20 & 94.51 & \textbf{90.09} & 89.25 \\
        +pos+type & \textbf{55.35} & \textbf{50.65} & \textbf{92.97} & \textbf{81.29} & 50.58 & \textbf{25.74} & \textbf{94.54} & 90.04 & 88.92 \\
        +arc+type & 53.04 & 48.96 & 92.68 & 80.40 & 48.49 & 26.54 & 94.29 & 89.84 & 90.86 \\
        +seg+arc+type & 51.74 & 48.83 & 92.49 & 80.29 & 50.04 & 26.74 & 93.87 & 89.19 & 86.34 \\
        +seg+pos+type & 52.33 & 49.05 & 92.68 & 80.60 & 50.01 & 26.72 & 93.48 & 89.28 & 89.46 \\
        +seg+pos+arc & 53.97 & 49.35 & 92.7 & 80.42 & 48.58 & 26.90 & 93.91 & 89.52 & 89.36 \\
        +pos+arc+type & 51.26 & 47.48 & 92.37 & 79.65 & 48.44 & 27.35 & 93.71 & 89.04 & 91.40 \\
        +all & 52.29 & 47.48 & 92.60 & 79.85 & 46.40 & 27.37 & 93.62 & 89.18 & 90.11 \\
        \hline
    \end{tabular}
    }
    \caption{Experimental results on bert-base-japanese-char.}
    \label{table9}
\end{table*}

\end{document}